%% file: sample-sigconf.tex
\documentclass[sigconf]{acmart}

\AtBeginDocument{%
  }

\setcopyright{acmlicensed}
\copyrightyear{2018}
\acmYear{2018}
\acmDOI{XXXXXXX.XXXXXXX}

\acmConference[Conference acronym 'XX]{Make sure to enter the correct
  conference title from your rights confirmation emai}{June 03--05,
  2018}{Woodstock, NY}
\acmISBN{978-1-4503-XXXX-X/18/06}

\usepackage{multirow}
\usepackage{amsmath}
\usepackage{bbold} 
\usepackage{algorithmic} 
\usepackage{algorithm}
\usepackage{booktabs}
\usepackage{bbm}

\begin{document}

\title{In-Context Learning with Noisy Labels}

\author{Junyong Kang}
\email{jykang@kaist.ac.kr}
\affiliation{%
  \institution{KAIST}
  \city{Seoul}
  \country{South Korea}
}

\author{Donghyun Son}
\email{happydh1@snu.ac.kr}
\affiliation{%
  \institution{Seoul National University}
  \city{Seoul}
  \country{South Korea}
}

\author{Hwanjun Song}
\email{songhwanjun@kaist.ac.kr}
\affiliation{%
  \institution{KAIST}
  \city{Deajeon}
  \country{South Korea}
}

\author{Buru Chang$^{*}$}
\thanks{$^{*}$ The corresponding author.}
\email{buru@sogang.ac.kr}
\affiliation{%
  \institution{Sogang University}
  \city{Seoul}
  \country{South Korea}
}

\renewcommand{\shortauthors}{Kang et al.}

\input{Sections/0_abstract}



\maketitle

\input{Sections/1_introduction}
\input{Sections/2_problem_formulation}
\input{Sections/3_method}
\input{Sections/4_experiments}

\input{Sections/5_conclusion}

\bibliographystyle{ACM-Reference-Format}
\bibliography{sample-base}


\end{document}

%% file: Sections/0_abstract.tex
\begin{abstract}\label{sec:0_abstract}
In-context learning refers to the emerging ability of large language models (LLMs) to perform a target task without additional training, utilizing demonstrations of the task. 
Recent studies aim to enhance in-context learning performance by selecting more useful demonstrations. 
However, they overlook the presence of inevitable noisy labels in task demonstrations that arise during the labeling process in the real-world. 
In this paper, we propose a new task, \textit{in-context learning with noisy labels}, which aims to solve real-world problems for in-context learning where labels in task demonstrations would be corrupted.
Moreover, we propose a new method and baseline methods for the new task, inspired by studies in learning with noisy labels.
Through experiments, we demonstrate that our proposed method can serve as a safeguard against performance degradation in in-context learning caused by noisy labels.
\end{abstract}

%% file: Sections/1_introduction.tex
\section{Introduction}\label{sec:1_introduction}
\noindent
\textit{Have you ever doubted whether the labels in your exemplars could be incorrect?}

\textit{In-context learning}~\cite{brown2020language,xie2021explanation} is an emerging property of Large Language Models (LLMs) trained on vast amounts of training data. 
Without the need for additional fine-tuning on the target task, LLMs can perform the target task by taking task descriptions and a few demonstrations (exemplars) relevant to the task.

Recently, many studies have been proposed to enhance the performance of in-context learning.
These studies aim to select more suitable demonstrations from the predefined retrieval set for a given test data~\cite{rubin2022learning,hongjin2022selective,ye2023compositional} or enable the utilization of more demonstrations by alleviating the limitation of input length in LLMs~\cite{hao2022structured,cai2023scaling}.
However, these studies overlook the real-world scenarios that the retrieved demonstrations would include \textit{noisy labels}.

\input{Figures/1_motivation}
In real-world labeled datasets, such as hate speech detection datasets~\cite{badjatiya2017deep}, corrupted labels often emerge due to the subjective nature of annotators and label ambiguity~\cite{natarajan2013learning}.
Given attempts to scale up the usage of demonstrations to the order of 1,000~\cite{hao2022structured}, prompts unavoidably include noisy labels.
In addition to previous works \cite{min2022rethink,wang2023investigating}, our experimental results in Figure~\ref{fig:1_motivation} show that these noisy labels not only severely degrade the performance of in-context learning but also reduce the stability of the predictive results (See details in $\S4.4$).
This observation underscores the need to handle noisy labels, as LLMs actually learn tasks from exemplars through in-context learning~\cite{von2023transformers,akyurek2022learning}, and such labels can interfere with this learning process.

To address this issue, we introduce a novel task, named \textit{"in-context learning with noisy labels,"} by establishing a connection between the existing \textit{"learning with noisy labels"} problem~\cite{natarajan2013learning,kye2022learning} and \textit{"in-context learning"} problem in LLMs.
The former problem aims to build a robust model from the corrupted dataset having noisy labels.
The new task assumes that a certain proportion of labels in a retrieval set are corrupted and evaluates the performance of in-context learning accordingly.

As the first comprehensive study on the new task, we present baseline methods for the new task. 
These baseline methods are adapted versions of representative methods proposed in existing research on learning with noisy labels~\cite{liu2015classification,yi2019probabilistic,zheng2021meta}, tailored to the in-context learning scenario.
Furthermore, we propose a new rectifying method that is more universally applicable to the in-context learning with noisy label setting.
Experiments on classification tasks demonstrate the importance of handling noisy labels in the in-context learning setting, and the proposed method shows its ability to serve as a safeguard against performance degradation caused by noisy labels.

%% file: Figures/1_motivation.tex
\begin{figure}[t]
    \centering
    \includegraphics[width=\columnwidth]{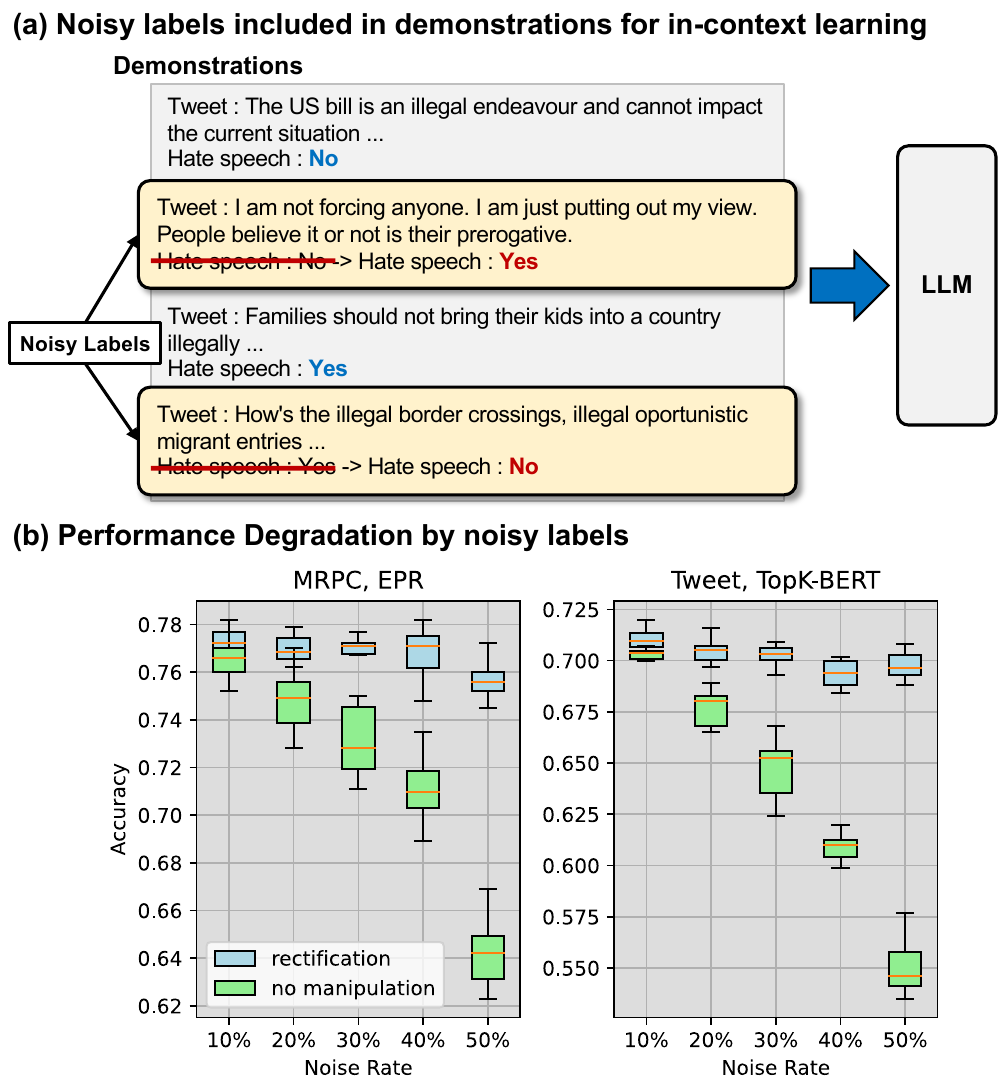}
    \caption{(a) The Tweet dataset~\cite{badjatiya2017deep} contains \textit{noisy labels}. Thus, the demonstrations collected from the dataset would include noisy labels. (b) These noisy labels degrade the performance of in-context learning and decrease the stability.}
    \label{fig:1_motivation}
\end{figure}

%% file: Sections/2_problem_formulation.tex
\section{Problem Formulation}\label{sec:3_problem_formulation}
In this section, we formally present a new task, denoted as \textit{in-context learning with noisy labels}.
Firstly, we formulate the problem of in-context learning, then extend the problem to the new task to address the real-world scenarios where the demonstrations would include noisy labels.

\paragraph{In-context learning.}
In-context learning is a paradigm where the LLM performs a new task without additional training by leveraging its powerful knowledge.
For this purpose, the LLM predicts the target label $y$ along with the input query $x$ and $n$-ordered demonstrations for the target task.
Following the previous studies~\cite{rubin2022learning,ye2023compositional}, we retrieve the most relevant demonstrations $[(x_1,y_1),\cdots,(x_n,y_n)]$ from the pre-defined retrieval set based on the input query $x$.
Subsequently, the demonstrations are concatenated to the input query $x$ as follows $[x_1,y_1,\cdots,x_n,y_n,x]$, and then it is fed to the frozen LLM to compute the probability distribution. 
Note that we focus on text classification tasks where the target label $y$ is in a set of $m$ candidate labels $\{c_1,\cdots,c_m\}$ to simplify our problem.
Thus, we decode the prediction $y$ by comparing the negative log-likelihood of the candidate labels and choose the minimal one.

\paragraph{In-context learning with noisy labels.}
In real-world labeled datasets, labels would be corrupted due to the nature of labeling process~\cite{natarajan2013learning}.
Therefore, we present a new task, \textit{in-context learning with noisy labels}, by extending the problem of in-context learning to consider the real-world scenarios.

We assume that the pre-defined retrieval set $\mathcal{D}$ is corrupted by various noise rates $r_i \in [0,1]$, yielding noisy retrieval set.
Specifically, we select $r_i \times |\mathcal{D}|$ demonstrations and flip each of label $y$ to $\Bar{y} \in \{c_1, ..., c_m\}- \{y\}$ with uniform transition probability, following the previous study~\cite{natarajan2013learning}.
As in traditional learning with noisy labels problem~\cite{song2023learning,hendrycks2018using,kye2022learning}, we assume that a small size of clean dataset is accessible.
Such a small-sized clean set can be constructed at a relatively low cost and is effective in significantly building a robust model to noisy labels.
We sample a small clean subset $\mathcal{D}'$ and use it to make models more robust to label noise.



%% file: Sections/3_method.tex
\section{Method}\label{sec:4_method}
We present several methods to perform the new task making in-context learning more robust in noisy labels.
In contrast to the learning with noisy labels problem, a significant challenge in the new task is that parameters of the LLM must be kept frozen instead of updating them to perform the task. 
Hence, we focus on manipulating the labels of task demonstrations.  

\input{Figures/2_trainformat}
\subsection{Baseline Methods}\label{subsec:3_1_baseline_methods}
We present four baseline methods for standard classification tasks named \textit{correction}, \textit{weighting}, \textit{reordering}, and \textit{selection}.
In learning with noisy labels problem, methods such as loss adjustment and sample selection~\cite{song2023learning} manipulate the effect of a noisy example by loss reweighting and sampling strategy based on its confidence. 
Motivated from these approaches, our baseline methods utilize a classifier (e.g. fine-tuned BERT) trained on each target dataset to give auxiliary information to LLM using its output probabilities.

\paragraph{Correction} is the simplest method which aims to directly correct noisy labels. 
This approach overwrites all demonstration labels with the output decision of the classifier.

\paragraph{Weighting} puts additional information next to the each prompt label, showing its confidence value from the classifier. 
We verbalize the confidence value as either \textit{"high"} or \textit{"low"} rather than the real number, as classifiers can produce extremely skewed scores~\cite{guo2017calibration}.

\paragraph{Reordering} utilizes the confidence value to replace position of each demonstration, placing low-confidence demonstrations proceeding to high-confidence ones.
LLMs puts different importance to each demonstrations depending on its position, typically higher weights to last sentences~\cite{zhao21c,lu2022fantastically}.
Therefore, reordering can be interpreted as an implicit weighting by the order bias of LLMs.

\paragraph{Selection} also utilizes the confidence value to reduce the influences of noisy labels by discarding demonstrations with low confidence labels in the prompt. 
As an example, we take the threshold $\theta = 0.3$ for the experiments.

\input{Tables/1_main_result}

\subsection{Rectification}\label{subsec:3_2_rectifying_network}
We now propose our main method \textit{rectification}, which processes multiple noisy demonstrations at a time.
As shown in Figure~\ref{fig:2_train_format}, rectification receives a sequence of noisy demonstrations and outputs a sequence of corrected labels.
To achieve this, we fine-tune a pre-trained generative model (e.g., GPT-2~\cite{Radford2019LanguageMA}) using negative log likelihood loss on the label tokens.
Unlike baseline methods that perform classification independently for each demonstration, rectification gets a sequence of demonstrations as an input.
Hence, it can reference all retrieved demonstrations and utilize them for noisy label rectification.
In $\S$4.4, we demonstrate the effectiveness of our design choice in improving model performance.

%% file: Figures/2_trainformat.tex
\begin{figure}[t]
    \includegraphics[width=\columnwidth]{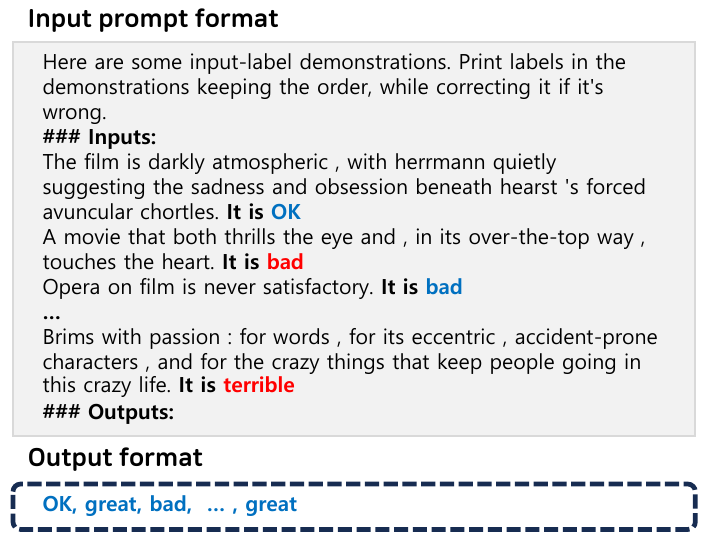}
    \caption{An example of the prompt format of the rectification method. The examples are collected from SST-5.}
    \label{fig:2_train_format}
\end{figure}

%% file: Tables/1_main_result.tex
\begin{table*}
\caption{Experimental results on GPT2-Neo (2.7B), Llama2-7B, and Mistral-7B models.}
\resizebox{\textwidth}{!}{%
\begin{tabular}{c|c|c|clcccc|clcccc|clcccc}
\toprule
\multirow{2}{*}{Model}&\multirow{2}{*}{Retriever} &
  \multirow{2}{*}{Method} &
  \multicolumn{6}{c|}{MRPC} &
  \multicolumn{6}{c|}{SST-5} &
  \multicolumn{6}{c}{Tweet} \\
 & & &0 &\multicolumn{1}{c}{0.1} &
  0.2 &
  0.3 &
  0.4 &
  0.5 &
  0 &
  \multicolumn{1}{c}{0.1} &
  0.2 &
  0.3 &
  0.4 &
  0.5 &
  0 &
  \multicolumn{1}{c}{0.1} &
  0.2 &
  0.3 &
  0.4 &
  0.5 \\ \midrule
\multirow{12}{*}{GPT2-Neo} &\multirow{6}{*}{EPR} &
  w/o manipulation &
  78.4 &
  76.0 &
  75.5 &
  67.9 &
  47.8 &
  51.5 &
  46.9 &
  44.1 &
  41.6 &
  38.5 &
  36.0 &
  31.8 &
  \textbf{58.4} &
  \textbf{59.7} &
  55.5 &
  54.7 &
  53.6 &
  47.4 \\ \cline{3-21} 
 & &
  Correction &
  \multicolumn{6}{c|}{72.1} &
  \multicolumn{6}{c|}{38.9} &
  \multicolumn{6}{c}{57.3} \\  
 & &
  Weighting &
  78.2 &
  77.2 &
  74.8 &
  70.8 &
  53.4 &
  58.3 &
  \textbf{48.5} &
  46.8 &
  44.5 &
  42.5 &
  39.2 &
  35.9 &
  58.1 &
  61.4 &
  57.2 &
  55.4 &
  54.0 &
  46.8 \\
 & &
  Reordering &
  \textbf{79.4} &
  68.1 &
  62.0 &
  49.0 &
  35.0 &
  36.8 &
  46.6 &
  46.3 &
  \textbf{45.3} &
  42.5 &
  38.5 &
  35.9 &
  57.7 &
  58.3 &
  57.5 &
  58.7 &
  58.0 &
  55.4 \\
 &&
  Selection &
  56.4 &
  49.3 &
  43.6 &
  40.7 &
  35.0 &
  36.3 &
  45.6 &
  44.7 &
  44.4 &
  43.2 &
  43.2 &
  38.5 &
  57.3 &
  57.3 &
  57.3 &
  57.3 &
  57.3 &
  56.4 \\ \cline{3-21} 
 &&
  Rectification &
  78.4 &
  \textbf{77.7} &
  \textbf{77.2} &
  \textbf{77.2} &
  \textbf{75.5} &
  \textbf{76.2} &
  44.7 &
  \textbf{44.8} &
  44.4 &
  \textbf{45.3} &
  \textbf{45.5} &
  \textbf{44.5} &
  58.3 &
  58.3 &
  \textbf{58.3} &
  \textbf{58.4} &
  \textbf{58.5} &
  \textbf{58.7} \\ \cline{2-21}
&\multirow{6}{*}{\begin{tabular}[c]{@{}c@{}}TopK-\\ BERT\end{tabular}} &
  w/o manipulation &
  70.3 &
  65.4 &
  62.5 &
  61.0 &
  53.4 &
  52.5 &
  35.6 &
  33.6 &
  31.7 &
  33.0 &
  29.0 &
  26.4 &
  \textbf{65.7} &
  63.3 &
  60.1 &
  58.8 &
  52.8 &
  50.3 \\ \cline{3-21} 
 &&
  Correction &
  \multicolumn{6}{c|}{67.0} &
  \multicolumn{6}{c|}{33.8} &
  \multicolumn{6}{c}{57.3} \\ 
 &&
  Weighting &
  \textbf{71.6} &
  69.1 &
  67.2 &
  65.4 &
  56.4 &
  56.4 &
  \textbf{36.9} &
  36.5 &
  35.5 &
  32.7 &
  33.9 &
  30.8 &
  65.3 &
  63.8 &
  59.6 &
  57.0 &
  50.2 &
  48.7 \\
 & &
  Reordering &
  52.7 &
  46.6 &
  42.2 &
  40.2 &
  34.6 &
  34.3 &
  36.4 &
  36.1 &
  35.1 &
  33.8 &
  31.9 &
  29.1 &
  69.7 &
  66.1 &
  62.7 &
  60.4 &
  56.4 &
  58.5 \\
 & &
  Selection &
  44.6 &
  41.4 &
  40.0 &
  35.5 &
  33.8 &
  32.8 &
  34.3 &
  33.8 &
  34.9 &
  33.2 &
  32.4 &
  32.9 &
  61.5 &
  60.6 &
  59.9 &
  59.5 &
  56.3 &
  57.7 \\ \cline{3-21}
 & &
  Rectification &
  71.1 &
  \textbf{71.1} &
  \textbf{70.3} &
  \textbf{71.3} &
  \textbf{68.2} &
  \textbf{67.4} &
  36.6 &
  \textbf{36.8} &
  \textbf{36.0} &
  \textbf{35.6} &
  \textbf{36.0} &
  \textbf{36.2} &
  64.4 &
  \textbf{64.2} &
  \textbf{63.7} &
  \textbf{63.1} &
  \textbf{63.5} &
  \textbf{64.0} \\ \midrule
  \multirow{6}{*}{Llama2-7B} &
  \multirow{3}{*}{EPR} &
  w/o manipulation &
  77.7 &
  75.5 &
  76.5 &
  74.3 &
  62.3 &
  62.3 &
  50.2 &
  49.3 &
  \textbf{51.0} &
  47.6 &
  45.3 &
  45.4 &
  \textbf{59.5} &
  \textbf{61.5} &
  58.6 &
  60.0 &
  57.5 &
  52.9 \\ \cline{3-21} 
 &
   &
  Correction &
  \multicolumn{6}{c|}{72.5} &
  \multicolumn{6}{c|}{43.7} &
  \multicolumn{6}{c}{57.3} \\ \cline{3-21} 
 &
   &
  Rectification &
  \textbf{78.2} &
  \textbf{78.4} &
  \textbf{78.4} &
  \textbf{78.4} &
  \textbf{76.7} &
  \textbf{75.8} &
  \textbf{50.3} &
  \textbf{50.5} &
  50.3 &
  \textbf{50.2} &
  \textbf{50.6} &
  \textbf{49.4} &
  59.9 &
  59.8 &
  \textbf{59.9} &
  \textbf{59.6} &
  \textbf{60.3} &
  \textbf{59.4} \\ \cline{2-21} 
 &
  \multirow{3}{*}{\begin{tabular}[c]{@{}c@{}}TopK-\\ BERT\end{tabular}} &
  w/o manipulation &
  70.3 &
  70.6 &
  69.1 &
  66.9 &
  59.1 &
  60.0 &
  49.8 &
  47.2 &
  49.0 &
  47.3 &
  43.5 &
  40.4 &
  71.4 &
  69.7 &
  67.5 &
  64.7 &
  56.9 &
  55.8 \\ \cline{3-21} 
 &
   &
  Correction &
  \multicolumn{6}{c|}{69.6} &
  \multicolumn{6}{c|}{46.3} &
  \multicolumn{6}{c}{57.4} \\ \cline{3-21} 
 &
   &
  Rectification &
  \textbf{73.5} &
  \textbf{71.6} &
  \textbf{72.3} &
  \textbf{70.6} &
  \textbf{71.3} &
  \textbf{71.1} &
  \textbf{51.2} &
  \textbf{50.8} &
  \textbf{51.1} &
  \textbf{50.2} &
  \textbf{50.6} &
  \textbf{49.4} &
  \textbf{71.7} &
  \textbf{71.7} &
  \textbf{71.3} &
  \textbf{69.7} &
  \textbf{71.3} &
  \textbf{70.3} \\ \midrule
\multirow{6}{*}{Mistral-7B} &
  \multirow{3}{*}{EPR} &
  w/o manipulation &
  77.7 &
  77.0 &
  76.5 &
  76.7 &
  61.3 &
  64.0 &
  \textbf{51.9} &
  \textbf{50.1} &
  48.5 &
  45.5 &
  44.3 &
  40.0 &
  \textbf{66.3} &
  \textbf{68.2} &
  63.3 &
  65.1 &
  61.8 &
  55.8 \\ \cline{3-21} 
 &
   &
  Correction &
  \multicolumn{6}{c|}{74.5} &
  \multicolumn{6}{c|}{44.8} &
  \multicolumn{6}{c}{57.3} \\ \cline{3-21} 
 &
   &
  Rectification &
  \textbf{78.4} &
  \textbf{77.9} &
  \textbf{77.9} &
  \textbf{78.9} &
  \textbf{77.0} &
  \textbf{77.0} &
  49.4 &
  49.9 &
  \textbf{49.5} &
  \textbf{49.9} &
  \textbf{49.8} &
  \textbf{49.8} &
  66.0 &
  65.8 &
  \textbf{65.8} &
  \textbf{65.8} &
  \textbf{65.8} &
  \textbf{65.8} \\ \cline{2-21} 
 &
  \multirow{3}{*}{\begin{tabular}[c]{@{}c@{}}TopK-\\ BERT\end{tabular}} &
  w/o manipulation &
  72.3 &
  72.1 &
  70.3 &
  67.9 &
  62.3 &
  62.3 &
  47.7 &
  46.4 &
  44.8 &
  42.7 &
  40.3 &
  39.2 &
  \textbf{74.6} &
  \textbf{71.5} &
  67.5 &
  65.9 &
  57.5 &
  57.5 \\ \cline{3-21} 
 &
   &
  Correction &
  \multicolumn{6}{c|}{71.6} &
  \multicolumn{6}{c|}{43.6} &
  \multicolumn{6}{c}{57.2} \\ \cline{3-21} 
 &
   &
  Rectification &
  \textbf{73.0} &
  \textbf{72.3} &
  \textbf{72.5} &
  \textbf{71.8} &
  \textbf{71.8} &
  \textbf{71.6} &
  \textbf{49.0} &
  \textbf{48.7} &
  \textbf{49.7} &
  \textbf{49.1} &
  \textbf{50.2} &
  \textbf{48.4} &
  72.1 &
  \textbf{71.5} &
  \textbf{71.4} &
  \textbf{70.2} &
  \textbf{70.3} &
  \textbf{69.1} \\ \bottomrule
\end{tabular}%
}
\label{tab:1_main_result}

\end{table*}

%% file: Sections/4_experiments.tex
\input{Tables/data_stat}
\section{Experiments}\label{sec:4_experiments}
We conduct experiments on three datasets: MRPC (paraphrase detection) ~\cite{dolan2005mrpc}, SST-5 (sentiment analysis) ~\cite{socher2013sst5} and Tweet hate speech detection~\cite{basile-etal-2019-semeval}.
The statistics of the datasets are summarized in Table~\ref{tab:statistics}.
We employ the training set to build the corrupted retrieval set $\mathcal{D}$ and sample a clean subset $\mathcal{D}'$ corresponding to 10\% of the training set.
Then, we evaluate the performance of our methods for \textit{in-context learning with noisy labels} on the validation set.

\subsection{Implementation Details}\label{subsec:4_1_impltdetails}
We adopt multiple LLMs of varying sizes, including GPT2-Neo 2.7B~\cite{Black2021gpt}, Llama2-7B~\cite{Touvron2023Llama2O}, and Mistral-7B~\cite{jiang2023mistral}, as inference LLMs, and use 10 task demonstrations for both training and inference.
For baseline methods, we fine-tune a BERT model~\cite{Devlin2019BERTPO} on the clean subset $\mathcal{D}'$, initialized with the \texttt{bert-base-uncased} checkpoint.
We also employ the oracle classifier, which is trained on clean retrieval set $\mathcal{D}$.
These classification models are fine-tuned for 30 epochs with batch size of 64 and learning rate of 5e-5.
For the rectification method, we fine-tune a pre-trained \texttt{gpt2-large}, one of pre-trained versions of GPT-2~\cite{Radford2019LanguageMA}.
To construct training dataset for the rectification method, we collect task demonstrations from the clean subset $\mathcal{D}'$ with the EPR retriever~\cite{rubin2022learning}, and then add label noise to the demonstrations by sampling $r \in \{0.1,0.2,0.3,0.4,0.5\}$.
We trained the model for 10 epochs with a batch size of 2 and a learning rate of 1e-4, employing LoRA~\cite{hu2022lora} for memory-efficient training.
We split the training set for these methods to create a validation set.

\subsection{Main results}\label{subsec:4_2_baseline_results}
Table~\ref{tab:1_main_result} shows the performance of our baseline methods and the rectification method.
We report the performance using the EPR retriever and the TopK-similarity retriever based on BERT embedding of demonstration inputs.
In the GPT2-Neo results, without any label manipulation, we observe that the performance of in-context learning degrades as the noise rate increases.
In contrast, the correction method acts as a robust baseline, maintaining consistent performance across all noise rates.
The weighting method shows some improvement compared to no manipulation, but fails to achieve consistent performance as the noise rate increases and sometimes performs worse in the Tweet.
The reordering and the selection method shows rapid performance degradation in MRPC due to the imperfect classifier.
In SST-5, the selection is relatively steady, and the reordering retains high accuracy at lower noise rates.
Although these methods successfully defends against performance drop in Tweet with the EPR retriever, we observe failures when using the TopK-BERT retriever.
On the other hand, the rectification method shows superior performance over baseline methods, effectively defending the performance against noisy labels.

For Llama2-7B and Mistral-7B, we report the performance of the correction and the rectification method.
These LLMs show better noise robustness at lower noise rates, Llama2-7B on SST-5 or Mistral-7B on Tweet with the EPR retriever for instance. 
However, their robustness is not consistent across datasets or retrievers.
For example, we found that the EPR mostly retrieves non-hate demonstrations for evaluation queries, which induce lower performance but higher robustness.
On the other hand, the TopK-BERT retrieves more balanced demonstrations, resulting in a clear drop in accuracy as the noise rate increases.
Regardless of these variations, the rectification method consistently maintains accuracy across noise rates and outperforms the correction method.

\subsection{Analysis}\label{subsec:4_3_results_analysis}

\input{Tables/5_stability}
\textbf{Stability.}
With our method, not only accuracy but also stability can be improved.
In Figure \ref{fig:1_motivation}, we visualize the in-context learning performance of the no manipulation version and the rectification method with 10 different random seeds. 
We also summarize the mean and standard deviation of performance for each dataset in Table \ref{tab:5_stability}, averaging the mean and the standard deviation across noise rates.
To solely measure the variability due to each noise rate, we retrieve demonstrations from $\mathcal{D}$ and then randomly corrupt them by rate $r_i$ in this experiment.
Overall, the rectification method shows smaller variance compared to no manipulation within each noise rate, which indicates that the stability is improved.

\input{Tables/3_rectification_performance}
\noindent\textbf{Rectification Accuracy.}
The performance of in-context learning with noisy labels depends on how well it rectifies the noisy labels. 
To measure the rectification performance, 
we define the \textit{rectification accuracy} as $\tau = \frac{1}{NK}\sum^{N}_{n=1}\sum^{K}_{k=1}\mathbbm{1}(y^k_n=\Tilde{y}^k_n)$, where $N,K$ denote the number of sets of demonstrations and the number of demonstrations in a set, $y^K_n$ and $\Tilde{y}^k_n$ denote $k$'th label and output of $n$'th set, respectively.
In Table \ref{tab:3_rect_perf}, we compare rectification accuracy between classification models and our rectification method. 
Specifically, we adopt the BERT classifier from previous sections and a classifier trained from \texttt{gpt2-large} to match model capacity. 
The rectification method achieves higher rectification accuracy compared to classifiers across all datasets, showing that it can effectively leverages the context of demonstrations in correcting labels.

\input{Tables/4_semi_nshot}
\noindent\textbf{Data efficiency.}
As observed in previous sections, the rectification method maintains the performance using only a small portion of the data.
To explore the data efficiency of this method, we train the oracle model (the rectification method trained on the full dataset $\mathcal{D}$), which we denote as \textit{Full.}
As shown in Table~\ref{tab:4_sample_efficiency}, our method reaches comparable performance to \textit{Full}, though it only uses 10\% of the total data.

\noindent\textbf{Why is the rectification method data-efficient?}
To investigate this question, we train the rectification method with fewer input demonstrations, specifically 2 and 5.
During the inference, these models are executed multiple times (e.g. 5 and 2) to rectify total 10 demonstrations.
Table~\ref{tab:4_sample_efficiency} shows that these method are less effective than our approach, where 10 demonstrations are given as an input.
From this observation, we conclude that the rectification method benefits from referring to the given context (demonstrations), which we believe to be crucial for the data efficiency.


%% file: Tables/data_stat.tex
\begin{table}[t!]
\caption{Dataset statistics and input formats.}
\resizebox{\columnwidth}{!}{%
\begin{tabular}{@{}cccc@{}}
\toprule
Dataset              & \# of train data  & \# of validation data &Format  \\ \midrule
MRPC& 3,668 & 408 &  \begin{tabular}[c]{@{}c@{}} \textit{\{sentence1\}} Can we say \\ "\textit{\{sentence2\}}"? \texttt{\{No, Yes\}}\end{tabular}\\
\midrule
SST-5 & 8,534   & 1,101 &
  \begin{tabular}[c]{@{}c@{}}\textit{\{question\}} It is \texttt{\{terrible,} \\ \texttt{bad,OK,good,great\}}\end{tabular} \\
  \midrule
Tweet & 9,000 & 1,000 &
  \begin{tabular}[c]{@{}c@{}}Tweet: \textit{\{question\}} \\ Hate: \texttt{\{No, Yes\}}\end{tabular} \\ \bottomrule
\end{tabular}%
}
\label{tab:statistics}
\end{table}

%% file: Tables/5_stability.tex
\begin{table}[]
\caption{Experimental results on stability. For MRPC and SST-5, we used the EPR retriever and the TopK-BERT for Tweet.}
\resizebox{\columnwidth}{!}{%
\begin{tabular}{@{}cccccccc@{}}
\toprule
\multirow{2}{*}{Model} & \multirow{2}{*}{Rectification} & \multicolumn{2}{c}{MRPC} & \multicolumn{2}{c}{SST-5} & \multicolumn{2}{c}{Tweet} \\ \cmidrule(l){3-8} 
          &                 & Acc.           & Std.          & Acc.           & Std.          & Acc.           & Std.          \\ \midrule
\multirow{2}{*}{GPT2-Neo}  & No manipulation & 65.23          & 1.98          & 37.56          & 1.24          & 57.50          & 1.28          \\
          & Rectification   & \textbf{76.02} & \textbf{0.83} & \textbf{44.39} & \textbf{0.53} & \textbf{63.79} & \textbf{0.72} \\ \midrule
\multirow{2}{*}{Llama2-7B} & No manipulation & 71.99          & 1.35          & 47.21          & 0.99          & 63.71          & 1.01          \\
          & Rectification   & \textbf{77.15} & \textbf{0.80} & \textbf{49.49} & \textbf{0.64} & \textbf{70.21} & \textbf{0.65} \\ \bottomrule
\end{tabular}%
}
\label{tab:5_stability}
\end{table}

%% file: Tables/3_rectification_performance.tex
\begin{table}[t!]
\caption{The rectification accuracy comparison.  ($\ast$) symbol indicates the oracle classifier trained on the full training data.}
\centering
\small
\begin{tabular}{l|ccc}
\toprule
\multicolumn{1}{l}{Rectification Method} & \multicolumn{1}{c}{MRPC} & \multicolumn{1}{c}{SST-5} & \multicolumn{1}{c}{Tweet} \\ \midrule
\multicolumn{1}{l|}{BERT Classifier$\ast$} & 97.5& 59.8 &91.5\\
\multicolumn{1}{l|}{BERT Classifier} & 81.9& 49.6&91.6\\\midrule
\multicolumn{1}{l|}{GPT-2 Classifier$\ast$} & 80.9&49.2&87.5\\
\multicolumn{1}{l|}{GPT-2 Classifier} & 72.1&42.3&68.4\\\midrule
\multicolumn{1}{l|}{Rectification ($r$=0.1)} & 90.3 & 67.3 & 95.1\\               
\multicolumn{1}{l|}{Rectification ($r$=0.2)} & 88.6 & 66.3 & 95.1\\
\multicolumn{1}{l|}{Rectification ($r$=0.3)} & 86.1 & 65.2 & 94.8\\               
\multicolumn{1}{l|}{Rectification ($r$=0.4)} & 83.5 & 64.0 & 94.6\\ 
\multicolumn{1}{l|}{Rectification ($r$=0.5)} & 83.8 & 62.4 & 94.1\\               \bottomrule
\end{tabular}%

\label{tab:3_rect_perf}
\end{table}

%% file: Tables/4_semi_nshot.tex
\begin{table}[]
\caption{Experimental results with GPT2-Neo to investigate data efficiency. The results on MRPC with the EPR retriever are reported.}
\centering
\small
\begin{tabular}{@{}c|cccccc@{}}
\toprule
\multirow{2}{*}{Training Method} & \multicolumn{6}{c}{Noise rate} \\ \cmidrule(l){2-7} 
       & 0             & 0.1           & 0.2  & 0.3           & 0.4           & 0.5           \\ \midrule
Full   & 77.7          & \textbf{78.4} & \textbf{77.5} & 76.7          & \textbf{76.7} & \textbf{77.0} \\ \midrule
2-shot & 73.5          & 73.8          & 73.2 & 73.0          & 73.5          & 72.1          \\
5-shot & 73.3          & 73.8          & 73.3 & 71.3          & 70.4          & 70.1          \\
Ours   & \textbf{78.4} & 77.7          & 77.2 & \textbf{77.2} & 75.5          & 76.2          \\ \bottomrule
\end{tabular}
\label{tab:4_sample_efficiency}

\end{table}

%% file: Sections/5_conclusion.tex
\section{Conclusion}\label{sec:5_conclusion}
In this study, we propose a new task called in-context learning with noisy labels. 
Additionally, we introduce baseline methods capable of performing the new task and propose a novel method to address their limitations.
Given the recent surge in attempts to generate labeled data using large language models, noisy labels are inevitable.
Our research highlight a new research direction that must be addressed in the era of LLM.

%% file: sample-sigconf.bbl

\begin{thebibliography}{31}


\ifx \showCODEN    \undefined \def \showCODEN     #1{\unskip}     \fi
\ifx \showDOI      \undefined \def \showDOI       #1{#1}\fi
\ifx \showISBNx    \undefined \def \showISBNx     #1{\unskip}     \fi
\ifx \showISBNxiii \undefined \def \showISBNxiii  #1{\unskip}     \fi
\ifx \showISSN     \undefined \def \showISSN      #1{\unskip}     \fi
\ifx \showLCCN     \undefined \def \showLCCN      #1{\unskip}     \fi
\ifx \shownote     \undefined \def \shownote      #1{#1}          \fi
\ifx \showarticletitle \undefined \def \showarticletitle #1{#1}   \fi
\ifx \showURL      \undefined \def \showURL       {\relax}        \fi
\providecommand\bibfield[2]{#2}
\providecommand\bibinfo[2]{#2}
\providecommand\natexlab[1]{#1}
\providecommand\showeprint[2][]{arXiv:#2}

\bibitem[Tou(2023)]%
        {Touvron2023Llama2O}
 \bibinfo{year}{2023}\natexlab{}.
\newblock \showarticletitle{Llama 2: Open Foundation and Fine-Tuned Chat Models}.
\newblock \bibinfo{journal}{\emph{ArXiv}}  \bibinfo{volume}{abs/2307.09288} (\bibinfo{year}{2023}).
\newblock


\bibitem[Aky{\"u}rek et~al\mbox{.}(2022)]%
        {akyurek2022learning}
\bibfield{author}{\bibinfo{person}{Ekin Aky{\"u}rek}, \bibinfo{person}{Dale Schuurmans}, \bibinfo{person}{Jacob Andreas}, \bibinfo{person}{Tengyu Ma}, {and} \bibinfo{person}{Denny Zhou}.} \bibinfo{year}{2022}\natexlab{}.
\newblock \showarticletitle{What learning algorithm is in-context learning? Investigations with linear models}. In \bibinfo{booktitle}{\emph{The Eleventh International Conference on Learning Representations}}.
\newblock


\bibitem[Badjatiya et~al\mbox{.}(2017)]%
        {badjatiya2017deep}
\bibfield{author}{\bibinfo{person}{Pinkesh Badjatiya}, \bibinfo{person}{Shashank Gupta}, \bibinfo{person}{Manish Gupta}, {and} \bibinfo{person}{Vasudeva Varma}.} \bibinfo{year}{2017}\natexlab{}.
\newblock \showarticletitle{Deep learning for hate speech detection in tweets}. In \bibinfo{booktitle}{\emph{Proceedings of the 26th international conference on World Wide Web companion}}. \bibinfo{pages}{759--760}.
\newblock


\bibitem[Basile et~al\mbox{.}(2019)]%
        {basile-etal-2019-semeval}
\bibfield{author}{\bibinfo{person}{Valerio Basile}, \bibinfo{person}{Cristina Bosco}, \bibinfo{person}{Elisabetta Fersini}, \bibinfo{person}{Debora Nozza}, \bibinfo{person}{Viviana Patti}, \bibinfo{person}{Francisco~Manuel Rangel~Pardo}, \bibinfo{person}{Paolo Rosso}, {and} \bibinfo{person}{Manuela Sanguinetti}.} \bibinfo{year}{2019}\natexlab{}.
\newblock \showarticletitle{{S}em{E}val-2019 Task 5: Multilingual Detection of Hate Speech Against Immigrants and Women in {T}witter}. In \bibinfo{booktitle}{\emph{Proceedings of the 13th International Workshop on Semantic Evaluation}}. \bibinfo{publisher}{Association for Computational Linguistics}, \bibinfo{address}{Minneapolis, Minnesota, USA}, \bibinfo{pages}{54--63}.
\newblock
\urldef\tempurl%
\url{https://doi.org/10.18653/v1/S19-2007}
\showDOI{\tempurl}


\bibitem[Black et~al\mbox{.}(2021)]%
        {Black2021gpt}
\bibfield{author}{\bibinfo{person}{Sid Black}, \bibinfo{person}{Leo Gao}, \bibinfo{person}{Phil Wang}, \bibinfo{person}{Connor Leahy}, {and} \bibinfo{person}{Stella Biderman}.} \bibinfo{year}{2021}\natexlab{}.
\newblock \bibinfo{booktitle}{\emph{{GPT-Neo: Large Scale Autoregressive Language Modeling with Mesh-Tensorflow}}}.
\newblock
\urldef\tempurl%
\url{https://doi.org/10.5281/zenodo.5297715}
\showDOI{\tempurl}


\bibitem[Brown et~al\mbox{.}(2020)]%
        {brown2020language}
\bibfield{author}{\bibinfo{person}{Tom Brown}, \bibinfo{person}{Benjamin Mann}, \bibinfo{person}{Nick Ryder}, \bibinfo{person}{Melanie Subbiah}, \bibinfo{person}{Jared~D Kaplan}, \bibinfo{person}{Prafulla Dhariwal}, \bibinfo{person}{Arvind Neelakantan}, \bibinfo{person}{Pranav Shyam}, \bibinfo{person}{Girish Sastry}, \bibinfo{person}{Amanda Askell}, {et~al\mbox{.}}} \bibinfo{year}{2020}\natexlab{}.
\newblock \showarticletitle{Language models are few-shot learners}.
\newblock \bibinfo{journal}{\emph{Advances in neural information processing systems}}  \bibinfo{volume}{33} (\bibinfo{year}{2020}), \bibinfo{pages}{1877--1901}.
\newblock


\bibitem[Cai et~al\mbox{.}(2023)]%
        {cai2023scaling}
\bibfield{author}{\bibinfo{person}{Tianle Cai}, \bibinfo{person}{Kaixuan Huang}, \bibinfo{person}{Jason~D Lee}, {and} \bibinfo{person}{Mengdi Wang}.} \bibinfo{year}{2023}\natexlab{}.
\newblock \showarticletitle{Scaling In-Context Demonstrations with Structured Attention}. In \bibinfo{booktitle}{\emph{Workshop on Efficient Systems for Foundation Models@ ICML2023}}.
\newblock


\bibitem[Devlin et~al\mbox{.}(2019)]%
        {Devlin2019BERTPO}
\bibfield{author}{\bibinfo{person}{Jacob Devlin}, \bibinfo{person}{Ming-Wei Chang}, \bibinfo{person}{Kenton Lee}, {and} \bibinfo{person}{Kristina Toutanova}.} \bibinfo{year}{2019}\natexlab{}.
\newblock \showarticletitle{BERT: Pre-training of Deep Bidirectional Transformers for Language Understanding}. In \bibinfo{booktitle}{\emph{North American Chapter of the Association for Computational Linguistics}}.
\newblock


\bibitem[Dolan and Brockett(2005)]%
        {dolan2005mrpc}
\bibfield{author}{\bibinfo{person}{William~B. Dolan} {and} \bibinfo{person}{Chris Brockett}.} \bibinfo{year}{2005}\natexlab{}.
\newblock \showarticletitle{Automatically Constructing a Corpus of Sentential Paraphrases}. In \bibinfo{booktitle}{\emph{Proceedings of the Third International Workshop on Paraphrasing ({IWP}2005)}}.
\newblock
\urldef\tempurl%
\url{https://aclanthology.org/I05-5002}
\showURL{%
\tempurl}


\bibitem[Guo et~al\mbox{.}(2017)]%
        {guo2017calibration}
\bibfield{author}{\bibinfo{person}{Chuan Guo}, \bibinfo{person}{Geoff Pleiss}, \bibinfo{person}{Yu Sun}, {and} \bibinfo{person}{Kilian~Q. Weinberger}.} \bibinfo{year}{2017}\natexlab{}.
\newblock \showarticletitle{On calibration of modern neural networks}. In \bibinfo{booktitle}{\emph{Proceedings of the 34th International Conference on Machine Learning - Volume 70}} (Sydney, NSW, Australia) \emph{(\bibinfo{series}{ICML'17})}. \bibinfo{publisher}{JMLR.org}, \bibinfo{pages}{1321–1330}.
\newblock


\bibitem[Hao et~al\mbox{.}(2022)]%
        {hao2022structured}
\bibfield{author}{\bibinfo{person}{Yaru Hao}, \bibinfo{person}{Yutao Sun}, \bibinfo{person}{Li Dong}, \bibinfo{person}{Zhixiong Han}, \bibinfo{person}{Yuxian Gu}, {and} \bibinfo{person}{Furu Wei}.} \bibinfo{year}{2022}\natexlab{}.
\newblock \showarticletitle{Structured prompting: Scaling in-context learning to 1,000 examples}.
\newblock \bibinfo{journal}{\emph{arXiv preprint arXiv:2212.06713}} (\bibinfo{year}{2022}).
\newblock


\bibitem[Hendrycks et~al\mbox{.}(2018)]%
        {hendrycks2018using}
\bibfield{author}{\bibinfo{person}{Dan Hendrycks}, \bibinfo{person}{Mantas Mazeika}, \bibinfo{person}{Duncan Wilson}, {and} \bibinfo{person}{Kevin Gimpel}.} \bibinfo{year}{2018}\natexlab{}.
\newblock \showarticletitle{Using trusted data to train deep networks on labels corrupted by severe noise}.
\newblock \bibinfo{journal}{\emph{Advances in neural information processing systems}}  \bibinfo{volume}{31} (\bibinfo{year}{2018}).
\newblock


\bibitem[Hongjin et~al\mbox{.}(2022)]%
        {hongjin2022selective}
\bibfield{author}{\bibinfo{person}{SU Hongjin}, \bibinfo{person}{Jungo Kasai}, \bibinfo{person}{Chen~Henry Wu}, \bibinfo{person}{Weijia Shi}, \bibinfo{person}{Tianlu Wang}, \bibinfo{person}{Jiayi Xin}, \bibinfo{person}{Rui Zhang}, \bibinfo{person}{Mari Ostendorf}, \bibinfo{person}{Luke Zettlemoyer}, \bibinfo{person}{Noah~A Smith}, {et~al\mbox{.}}} \bibinfo{year}{2022}\natexlab{}.
\newblock \showarticletitle{Selective Annotation Makes Language Models Better Few-Shot Learners}. In \bibinfo{booktitle}{\emph{The Eleventh International Conference on Learning Representations}}.
\newblock


\bibitem[Hu et~al\mbox{.}(2022)]%
        {hu2022lora}
\bibfield{author}{\bibinfo{person}{Edward~J Hu}, \bibinfo{person}{Yelong Shen}, \bibinfo{person}{Phillip Wallis}, \bibinfo{person}{Zeyuan Allen-Zhu}, \bibinfo{person}{Yuanzhi Li}, \bibinfo{person}{Shean Wang}, \bibinfo{person}{Lu Wang}, {and} \bibinfo{person}{Weizhu Chen}.} \bibinfo{year}{2022}\natexlab{}.
\newblock \showarticletitle{Lo{RA}: Low-Rank Adaptation of Large Language Models}. In \bibinfo{booktitle}{\emph{International Conference on Learning Representations}}.
\newblock
\urldef\tempurl%
\url{https://openreview.net/forum?id=nZeVKeeFYf9}
\showURL{%
\tempurl}


\bibitem[Jiang et~al\mbox{.}(2023)]%
        {jiang2023mistral}
\bibfield{author}{\bibinfo{person}{Albert~Q Jiang}, \bibinfo{person}{Alexandre Sablayrolles}, \bibinfo{person}{Arthur Mensch}, \bibinfo{person}{Chris Bamford}, \bibinfo{person}{Devendra~Singh Chaplot}, \bibinfo{person}{Diego de~las Casas}, \bibinfo{person}{Florian Bressand}, \bibinfo{person}{Gianna Lengyel}, \bibinfo{person}{Guillaume Lample}, \bibinfo{person}{Lucile Saulnier}, {et~al\mbox{.}}} \bibinfo{year}{2023}\natexlab{}.
\newblock \showarticletitle{Mistral 7B}.
\newblock \bibinfo{journal}{\emph{arXiv preprint arXiv:2310.06825}} (\bibinfo{year}{2023}).
\newblock


\bibitem[Kye et~al\mbox{.}(2022)]%
        {kye2022learning}
\bibfield{author}{\bibinfo{person}{Seong~Min Kye}, \bibinfo{person}{Kwanghee Choi}, \bibinfo{person}{Joonyoung Yi}, {and} \bibinfo{person}{Buru Chang}.} \bibinfo{year}{2022}\natexlab{}.
\newblock \showarticletitle{Learning with noisy labels by efficient transition matrix estimation to combat label miscorrection}. In \bibinfo{booktitle}{\emph{European Conference on Computer Vision}}. Springer, \bibinfo{pages}{717--738}.
\newblock


\bibitem[Liu and Tao(2015)]%
        {liu2015classification}
\bibfield{author}{\bibinfo{person}{Tongliang Liu} {and} \bibinfo{person}{Dacheng Tao}.} \bibinfo{year}{2015}\natexlab{}.
\newblock \showarticletitle{Classification with noisy labels by importance reweighting}.
\newblock \bibinfo{journal}{\emph{IEEE Transactions on pattern analysis and machine intelligence}} \bibinfo{volume}{38}, \bibinfo{number}{3} (\bibinfo{year}{2015}), \bibinfo{pages}{447--461}.
\newblock


\bibitem[Lu et~al\mbox{.}(2022)]%
        {lu2022fantastically}
\bibfield{author}{\bibinfo{person}{Yao Lu}, \bibinfo{person}{Max Bartolo}, \bibinfo{person}{Alastair Moore}, \bibinfo{person}{Sebastian Riedel}, {and} \bibinfo{person}{Pontus Stenetorp}.} \bibinfo{year}{2022}\natexlab{}.
\newblock \showarticletitle{Fantastically Ordered Prompts and Where to Find Them: Overcoming Few-Shot Prompt Order Sensitivity}. In \bibinfo{booktitle}{\emph{Proceedings of the 60th Annual Meeting of the Association for Computational Linguistics (Volume 1: Long Papers)}}, \bibfield{editor}{\bibinfo{person}{Smaranda Muresan}, \bibinfo{person}{Preslav Nakov}, {and} \bibinfo{person}{Aline Villavicencio}} (Eds.). \bibinfo{publisher}{Association for Computational Linguistics}, \bibinfo{address}{Dublin, Ireland}, \bibinfo{pages}{8086--8098}.
\newblock
\urldef\tempurl%
\url{https://doi.org/10.18653/v1/2022.acl-long.556}
\showDOI{\tempurl}


\bibitem[Min et~al\mbox{.}(2022)]%
        {min2022rethink}
\bibfield{author}{\bibinfo{person}{Sewon Min}, \bibinfo{person}{Xinxi Lyu}, \bibinfo{person}{Ari Holtzman}, \bibinfo{person}{Mikel Artetxe}, \bibinfo{person}{Mike Lewis}, \bibinfo{person}{Hannaneh Hajishirzi}, {and} \bibinfo{person}{Luke Zettlemoyer}.} \bibinfo{year}{2022}\natexlab{}.
\newblock \showarticletitle{Rethinking the Role of Demonstrations: What Makes In-Context Learning Work?} \bibinfo{pages}{11048--11064}.
\newblock
\urldef\tempurl%
\url{https://doi.org/10.18653/v1/2022.emnlp-main.759}
\showDOI{\tempurl}


\bibitem[Natarajan et~al\mbox{.}(2013)]%
        {natarajan2013learning}
\bibfield{author}{\bibinfo{person}{Nagarajan Natarajan}, \bibinfo{person}{Inderjit~S Dhillon}, \bibinfo{person}{Pradeep~K Ravikumar}, {and} \bibinfo{person}{Ambuj Tewari}.} \bibinfo{year}{2013}\natexlab{}.
\newblock \showarticletitle{Learning with noisy labels}.
\newblock \bibinfo{journal}{\emph{Advances in neural information processing systems}}  \bibinfo{volume}{26} (\bibinfo{year}{2013}).
\newblock


\bibitem[Radford et~al\mbox{.}(2019)]%
        {Radford2019LanguageMA}
\bibfield{author}{\bibinfo{person}{Alec Radford}, \bibinfo{person}{Jeff Wu}, \bibinfo{person}{Rewon Child}, \bibinfo{person}{David Luan}, \bibinfo{person}{Dario Amodei}, {and} \bibinfo{person}{Ilya Sutskever}.} \bibinfo{year}{2019}\natexlab{}.
\newblock \showarticletitle{Language Models are Unsupervised Multitask Learners}.
\newblock


\bibitem[Rubin et~al\mbox{.}(2022)]%
        {rubin2022learning}
\bibfield{author}{\bibinfo{person}{Ohad Rubin}, \bibinfo{person}{Jonathan Herzig}, {and} \bibinfo{person}{Jonathan Berant}.} \bibinfo{year}{2022}\natexlab{}.
\newblock \showarticletitle{Learning To Retrieve Prompts for In-Context Learning}. In \bibinfo{booktitle}{\emph{Proceedings of the 2022 Conference of the North American Chapter of the Association for Computational Linguistics: Human Language Technologies}}. \bibinfo{pages}{2655--2671}.
\newblock


\bibitem[Socher et~al\mbox{.}(2013)]%
        {socher2013sst5}
\bibfield{author}{\bibinfo{person}{Richard Socher}, \bibinfo{person}{Alex Perelygin}, \bibinfo{person}{Jean Wu}, \bibinfo{person}{Jason Chuang}, \bibinfo{person}{Christopher~D. Manning}, \bibinfo{person}{Andrew Ng}, {and} \bibinfo{person}{Christopher Potts}.} \bibinfo{year}{2013}\natexlab{}.
\newblock \showarticletitle{Recursive Deep Models for Semantic Compositionality Over a Sentiment Treebank}. In \bibinfo{booktitle}{\emph{Proceedings of the 2013 Conference on Empirical Methods in Natural Language Processing}}, \bibfield{editor}{\bibinfo{person}{David Yarowsky}, \bibinfo{person}{Timothy Baldwin}, \bibinfo{person}{Anna Korhonen}, \bibinfo{person}{Karen Livescu}, {and} \bibinfo{person}{Steven Bethard}} (Eds.). \bibinfo{publisher}{Association for Computational Linguistics}, \bibinfo{address}{Seattle, Washington, USA}, \bibinfo{pages}{1631--1642}.
\newblock
\urldef\tempurl%
\url{https://aclanthology.org/D13-1170}
\showURL{%
\tempurl}


\bibitem[Song et~al\mbox{.}(2023)]%
        {song2023learning}
\bibfield{author}{\bibinfo{person}{Hwanjun Song}, \bibinfo{person}{Minseok Kim}, \bibinfo{person}{Dongmin Park}, \bibinfo{person}{Yooju Shin}, {and} \bibinfo{person}{Jae-Gil Lee}.} \bibinfo{year}{2023}\natexlab{}.
\newblock \showarticletitle{Learning From Noisy Labels With Deep Neural Networks: A Survey}.
\newblock \bibinfo{journal}{\emph{IEEE Transactions on Neural Networks and Learning Systems}} \bibinfo{volume}{34}, \bibinfo{number}{11} (\bibinfo{year}{2023}), \bibinfo{pages}{8135--8153}.
\newblock
\urldef\tempurl%
\url{https://doi.org/10.1109/TNNLS.2022.3152527}
\showDOI{\tempurl}


\bibitem[Von~Oswald et~al\mbox{.}(2023)]%
        {von2023transformers}
\bibfield{author}{\bibinfo{person}{Johannes Von~Oswald}, \bibinfo{person}{Eyvind Niklasson}, \bibinfo{person}{Ettore Randazzo}, \bibinfo{person}{Jo{\~a}o Sacramento}, \bibinfo{person}{Alexander Mordvintsev}, \bibinfo{person}{Andrey Zhmoginov}, {and} \bibinfo{person}{Max Vladymyrov}.} \bibinfo{year}{2023}\natexlab{}.
\newblock \showarticletitle{Transformers learn in-context by gradient descent}. In \bibinfo{booktitle}{\emph{International Conference on Machine Learning}}. PMLR, \bibinfo{pages}{35151--35174}.
\newblock


\bibitem[Wang et~al\mbox{.}(2023)]%
        {wang2023investigating}
\bibfield{author}{\bibinfo{person}{Xindi Wang}, \bibinfo{person}{Yufei Wang}, \bibinfo{person}{Can Xu}, \bibinfo{person}{Xiubo Geng}, \bibinfo{person}{Bowen Zhang}, \bibinfo{person}{Chongyang Tao}, \bibinfo{person}{Frank Rudzicz}, \bibinfo{person}{Robert~E Mercer}, {and} \bibinfo{person}{Daxin Jiang}.} \bibinfo{year}{2023}\natexlab{}.
\newblock \showarticletitle{Investigating the learning behaviour of in-context learning: a comparison with supervised learning}.
\newblock \bibinfo{journal}{\emph{arXiv preprint arXiv:2307.15411}} (\bibinfo{year}{2023}).
\newblock


\bibitem[Xie et~al\mbox{.}(2021)]%
        {xie2021explanation}
\bibfield{author}{\bibinfo{person}{Sang~Michael Xie}, \bibinfo{person}{Aditi Raghunathan}, \bibinfo{person}{Percy Liang}, {and} \bibinfo{person}{Tengyu Ma}.} \bibinfo{year}{2021}\natexlab{}.
\newblock \showarticletitle{An Explanation of In-context Learning as Implicit Bayesian Inference}. In \bibinfo{booktitle}{\emph{International Conference on Learning Representations}}.
\newblock


\bibitem[Ye et~al\mbox{.}(2023)]%
        {ye2023compositional}
\bibfield{author}{\bibinfo{person}{Jiacheng Ye}, \bibinfo{person}{Zhiyong Wu}, \bibinfo{person}{Jiangtao Feng}, \bibinfo{person}{Tao Yu}, {and} \bibinfo{person}{Lingpeng Kong}.} \bibinfo{year}{2023}\natexlab{}.
\newblock \showarticletitle{Compositional Exemplars for In-context Learning}. In \bibinfo{booktitle}{\emph{Proceedings of the 40th International Conference on Machine Learning}} \emph{(\bibinfo{series}{Proceedings of Machine Learning Research}, Vol.~\bibinfo{volume}{202})}, \bibfield{editor}{\bibinfo{person}{Andreas Krause}, \bibinfo{person}{Emma Brunskill}, \bibinfo{person}{Kyunghyun Cho}, \bibinfo{person}{Barbara Engelhardt}, \bibinfo{person}{Sivan Sabato}, {and} \bibinfo{person}{Jonathan Scarlett}} (Eds.). \bibinfo{publisher}{PMLR}, \bibinfo{pages}{39818--39833}.
\newblock
\urldef\tempurl%
\url{https://proceedings.mlr.press/v202/ye23c.html}
\showURL{%
\tempurl}


\bibitem[Yi and Wu(2019)]%
        {yi2019probabilistic}
\bibfield{author}{\bibinfo{person}{Kun Yi} {and} \bibinfo{person}{Jianxin Wu}.} \bibinfo{year}{2019}\natexlab{}.
\newblock \showarticletitle{Probabilistic end-to-end noise correction for learning with noisy labels}. In \bibinfo{booktitle}{\emph{Proceedings of the IEEE/CVF conference on computer vision and pattern recognition}}. \bibinfo{pages}{7017--7025}.
\newblock


\bibitem[Zhao et~al\mbox{.}(2021)]%
        {zhao21c}
\bibfield{author}{\bibinfo{person}{Zihao Zhao}, \bibinfo{person}{Eric Wallace}, \bibinfo{person}{Shi Feng}, \bibinfo{person}{Dan Klein}, {and} \bibinfo{person}{Sameer Singh}.} \bibinfo{year}{2021}\natexlab{}.
\newblock \showarticletitle{Calibrate Before Use: Improving Few-shot Performance of Language Models}. In \bibinfo{booktitle}{\emph{Proceedings of the 38th International Conference on Machine Learning}} \emph{(\bibinfo{series}{Proceedings of Machine Learning Research}, Vol.~\bibinfo{volume}{139})}, \bibfield{editor}{\bibinfo{person}{Marina Meila} {and} \bibinfo{person}{Tong Zhang}} (Eds.). \bibinfo{publisher}{PMLR}, \bibinfo{pages}{12697--12706}.
\newblock
\urldef\tempurl%
\url{https://proceedings.mlr.press/v139/zhao21c.html}
\showURL{%
\tempurl}


\bibitem[Zheng et~al\mbox{.}(2021)]%
        {zheng2021meta}
\bibfield{author}{\bibinfo{person}{Guoqing Zheng}, \bibinfo{person}{Ahmed~Hassan Awadallah}, {and} \bibinfo{person}{Susan Dumais}.} \bibinfo{year}{2021}\natexlab{}.
\newblock \showarticletitle{Meta label correction for noisy label learning}. In \bibinfo{booktitle}{\emph{Proceedings of the AAAI Conference on Artificial Intelligence}}, Vol.~\bibinfo{volume}{35}. \bibinfo{pages}{11053--11061}.
\newblock


\end{thebibliography}
